\documentclass[preprint,12pt]{elsarticle}

\usepackage{amssymb}

\journal{arXiv}

\makeatletter
\def\ps@pprintTitle{%
   \let\@oddfoot\@empty
   \let\@evenfoot\@empty
}
\makeatother

\begin{document}

\begin{frontmatter}



\title{PlateSegFL: A Privacy-Preserving License Plate Detection Using Federated Segmentation Learning}

\author[1]{Md. Shahriar Rahman Anuvab}
\author[1]{Mishkat Sultana}
\author[1]{Md. Atif Hossain}
\author[1]{Shashwata Das}
\author[1]{Suvarthi Chowdhury}
\author[1]{Rafeed Rahman}
\author[1]{\\Dibyo Fabian Dofadar}
\author[1]{Shahriar Rahman Rana}

\address[1]{Department of Computer Science and Engineering, Brac University, Dhaka, Bangladesh\\
\vspace{5mm}
Email: md.shahriar.rahman.anuvab@g.bracu.ac.bd, mishkat.sultana@g.bracu.ac.bd, md.atif.hossain@g.bracu.ac.bd, shashwata.das@g.bracu.ac.bd, suvarthi.chowdhury@g.bracu.ac.bd, rafeed.rahman@bracu.ac.bd, fabian.dofadar@bracu.ac.bd, shahriar.rahman@bracu.ac.bd}


\begin{abstract}
Automatic License Plate Recognition (ALPR) is an integral component of an intelligent transport system with extensive applications in secure transportation, vehicle-to-vehicle communication, stolen vehicles detection, traffic violations, and traffic flow management. The existing license plate detection system focuses on one-shot learners or pre-trained models that operate with a geometric bounding box, limiting the model’s performance. Furthermore, continuous video data streams uploaded to the central server result in network and complexity issues. To combat this, PlateSegFL was introduced, which implements U-Net-based segmentation along with Federated Learning (FL). U-Net is well-suited for multi-class image segmentation tasks because it can analyze a large number of classes and generate a pixel-level segmentation map for each class. Federated Learning is used to reduce the quantity of data required while safeguarding the user’s privacy. Different computing platforms, such as mobile phones, are able to collaborate on the development of a standard prediction model where it makes efficient use of one’s time; incorporates more diverse data; delivers projections in real-time; and requires no physical effort from the user; resulting around 95\% F1 score.
\end{abstract}

\begin{keyword}
Federated Learning, U-Net, Convolution Neural Network, License Plate Recognition
\end{keyword}

\end{frontmatter}




\section{Introduction}
\label{sec:Introduction}
The study aims to address the growing challenges related to the proliferation of automobiles and their role in violent crimes, emphasizes the necessity for efficient tracking and surveillance systems. This research takes a fresh approach to improving license plate identification skills while maintaining privacy by combining Federated Learning with U-Net, a strong segmentation technique. Such innovations are urgently needed due to the increasing crime rates, especially in car accidents \cite{song2023vehicle,shi2023license,zou2022license}. The study shows the U-Net-based Automatic license Plate Recognition (ALPR) design to deal with problems caused by irregular masking. It also talks about the problems with older methods, like bounding box segmentation. In order to decentralize model training without jeopardising user data, this research used Federated Learning to tackle privacy issues. The suggested methodology is an attempt to alleviate the problems associated with dataset collection, transmission overhead, and atypical masking that occurs during picture segmentation. The key achievements include the following: the use of U-Net for the identification of irregular mask segmentation; the use of Federated Learning to reduce transmission overhead while protecting privacy; and the use of Bangla and English OCR for the extraction of text from photos. In order to further the understanding of license plate identification and irregular masking, this research presents a thorough review and comparison with current models. In light of the rising number of vehicle-related crimes, this study's ultimate goal is to make a sizable contribution to the creation of a reliable and secure system for real-time number plate detection.

\setlength{\parindent}{5mm}


\section{Related Works}
\label{sec:Related Works}

In recent years, car plate detection has grabbed attention in the fields of computer vision and image processing. Among the many papers implying different approaches, they are mainly categorized into four groups, which are YOLO, YOLO Less, FL, and U-Net. About car plate detection, different studies have been explored with the approaches of YOLO V3 and YOLO V4 with deep learning methods. For instance, in \cite{setiyono2021number}, the authors have gained an 88\% accuracy after preprocessing. Additionally, \cite{riaz2020yolo} increased to 96\% using temporal redundancy. In the meantime, \cite{youssef2022real} gained 97.89\% accuracy using YOLO V3 in their work. Furthermore, the YOLO V4 system has a magnificent feature map and the ability to detect small license plates. Though YOLO V4 is still facing certain issues, including a low recall, some researchers proposed combining both YOLO V3 and YOLO V4 for car plate detection and identification, but there are some limitations like localization and memory allocation \cite{meena2021license,park2022all,raj2022license}. Above all, several studies \cite{gandhi2020yolo,al2022end} have shown significant progress in car plate detection but still there are rooms for improvement. 

To overcome the problems with YOLO V3 and YOLO V4, researchers \cite{ganjoo2022yolo, laroca2021efficient, chen2019automatic, pan2022research, castro2020license} introduced significant algorithms for improving dataset quality, segmentation, character recognition, and so on. CNN and YOLO are used in \cite{izidio2020embedded, angara2020license, calitz2020automated, tham2021iot, weihong2020research, jamtsho2020real} mostly for car plate detection and recognition. Along with that, Kernighan SVF and many other algorithms were used in the papers \cite{ ibitoye2020convolutional, li2022chinese, danilenko2020license, qin2020efficient}. These works emphasise region-specific challenges, while others explore data augmentation and integration \cite{li2022chinese,danilenko2020license,qin2020efficient,chen2020random,usmankhujaev2020korean}. All of these aim to enhance recognition accuracy and robustness, with the goal of creating more reliable systems capable of handling diverse license plate configurations and conditions.

Federated Learning (FL) is mostly used nowadays for car plate detection and recognition, addressing privacy and model deployment concerns. To enhance vehicle identification, \cite{chowdhury2020new} introduced FedLVR for privacy preservation. At the same time, \cite{tom2022car} enabled users to train models on their mobile devices, ensuring high accuracy, low communication costs, and privacy maintenance. Concurrent license plate identification from network cameras  achieved efficiency and better generalization after applying Federated Learning \cite{yu2022research}. 

A promising solution for robust car plate detection in challenging scenarios combines Federated Learning with U-Net segmentation. For effectiveness in image segmentation and irregular images, U-Net architecture is famous. In some recent works, \cite{zeng2023fedlvr,kong2021federated,xie2023asynchronous,rieke2020future}, U-Net enhances low-light license plate images, significantly improving recognition accuracy, especially in nighttime conditions. For a fast-moving autonomous vehicle monitoring system using a single-pass deep neural network, which integrates license plate detection and character recognition, \cite{rieke2020future} streamlined the process. Furthermore, \cite{yu2022research} presented an end-to-end license plate recognition approach that leverages U-Net for location and recognition, addressing challenging conditions such as slanted and small-scale license plates. These research works highlight the U-Net’s suitable handling of imperfect and irregular license plate images.
 
\section{Workflow}
\label{sec:Workflow}
Thorough model training starts with collecting datasets and manually masking them. Next, the U-Net segmentation model is used for training and testing, with Federated Learning taking care of privacy issues. Inscriptions from license plates are extracted, and the suggested model is shown to be superior to both traditional U-Net and YOLO in terms of computer vision and license plate recognition.

\begin{figure}[htbp]
\centerline{\includegraphics[width=\textwidth]{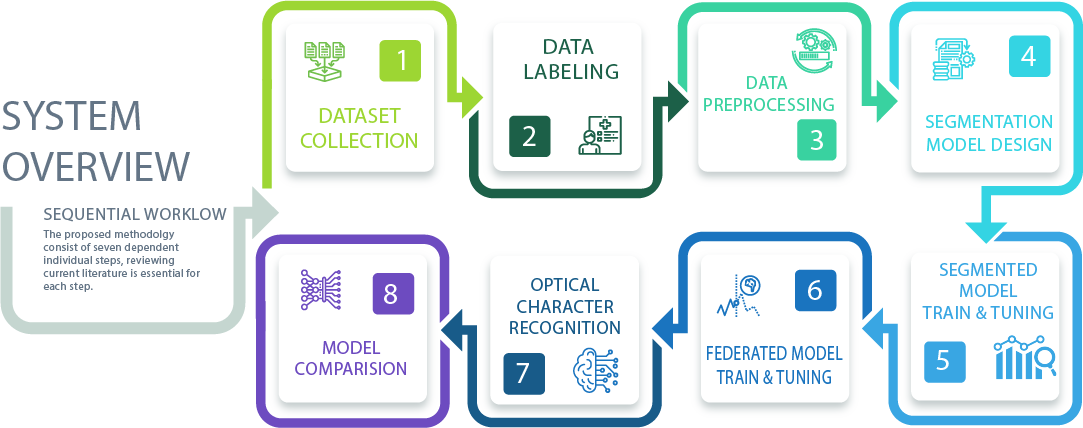}}
\vspace{7mm}
\caption{System Overview of Proposed Approach Using FL U-Net Model}
\label{fig:FU model}
\end{figure}

\section{Dataset}
\label{Dataset Details}

\begin{figure}[htpb!]
\centerline{\includegraphics[width=\textwidth]
{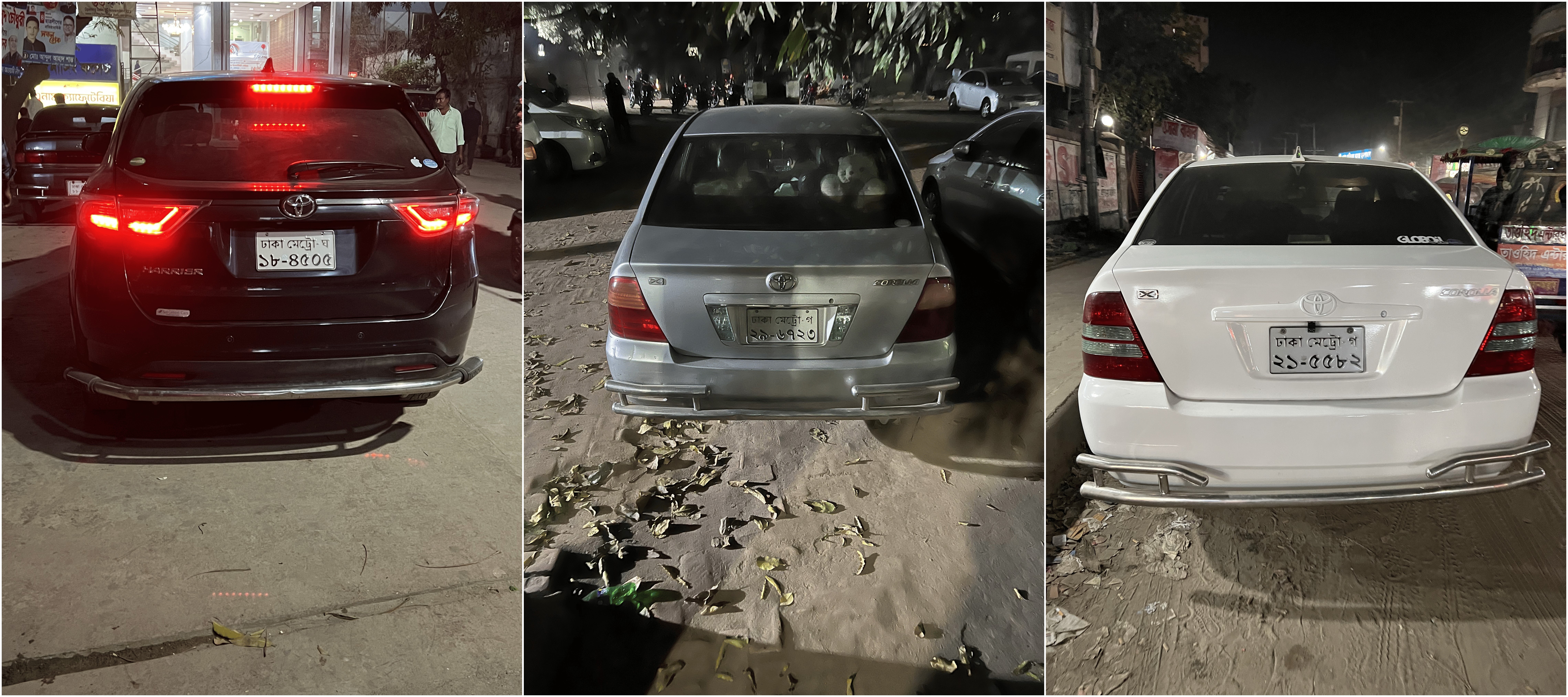}}
\vspace{3mm}
\caption{Bangladeshi Vehicle Images}
\label{fig:merge dataset}
\end{figure}

For the proposed approach, three datasets was collected in addition to firsthand collection of data. The dataset, “Car License Plate Detection" dataset \cite{makeml}, contains a total of 433 vehicle images. These images are from different countries in Europe and America. Similarly, in the Tunisian dataset “Labelled license Plate" \cite{ visualization-tools-for-tunisian-licensed-plates-dataset} which contains Tunisian license plates and the coordinates of each license plate, there are 707 images. Next, the “Bangla Dataset LPDB” \cite{ataher-shomee} is the first related dataset in Bangladesh and includes almost every kind of vehicle seen in Bangladesh roads. This includes 4590 images of Bangladeshi vehicles and license plates in Bangla. Lastly, some images were gathered manually by the authors from the streets of Bangladesh (Figure \ref{fig:merge dataset}), specifically from the roadways. A total of 373 still frames figure and three videos were taken in broad daylight and at night in diverse inclinations in broad day and night and in different weather conditions [Table \ref{tab:vehicle_percentages}]. The dataset was expanded from 6370 to 11,472 photos by data augmentation, which is essential for improving model accuracy and avoiding overfitting. Scaling, feature-wise standard normalization, width and height shift ranges, and feature-wise centering were among the augmentation techniques employed. For supervised learning to work, it is crucial to label both the input and output data accurately. Segmentation images were introduced to tackle the uneven masking issues in models like YOLO in this context. By manually masking license plates, accurate labelling was possible. This helped the model acquire the correct differences even in various situations. Two levels of masking were used to remove imperfections. The input mean and standard deviation were established using feature-wise centering and standard normalization. Data processing and model training were made more efficient with the help of width and height shift ranges, and resizing.

\begin{table}[h]
\centering
\begin{tabular}{|l|c|}
\hline
\textbf{Vehicle Type} & \textbf{Percentage} \\
\hline
Car & 65\% \\
\hline
Bike & 28\% \\
\hline
Others & 17\% \\
\hline
\end{tabular}
\caption{Vehicle Types and Percentages}
\label{tab:vehicle_percentages}
\end{table}

\section{Methodology}
\subsection{U-Net}
\label{U-Net}
\begin{figure}[htp!]
\vspace{10mm}
\centerline{\includegraphics[width=\textwidth]{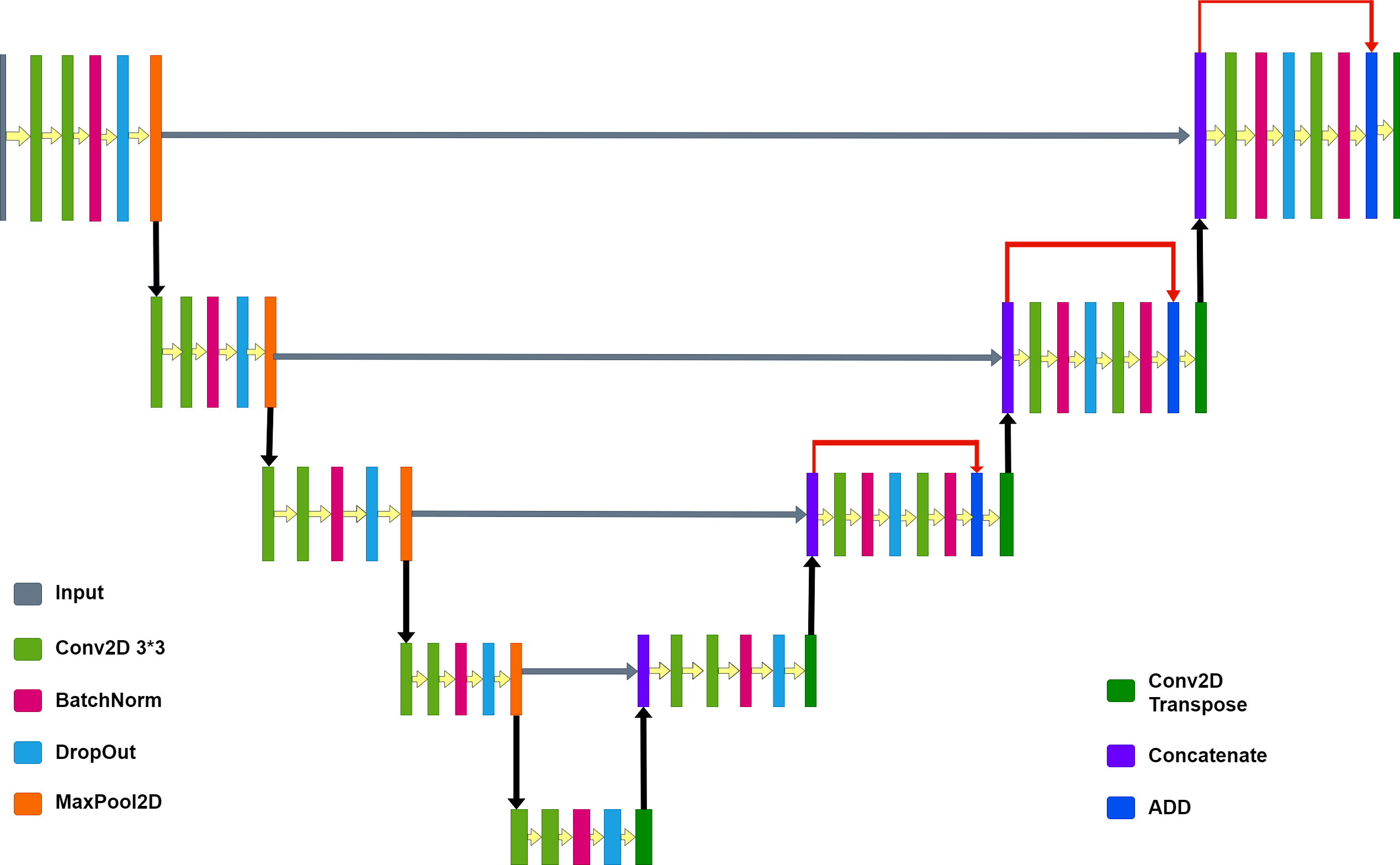}}
\vspace{7mm}
\caption{Proposed U-Net model}
\label{fig:pu}
\end{figure}

\begin{table}[ht!]
  \centering
 \begin{tabular}{|c|c|c|}
   \hline
    \textbf{Layer (type)}        & \textbf{Output Shape}        & \textbf{Parameters} \\ \hline
    img (InputLayer)              & (None, 192, 192, 3)         & 0                    \\ \hline
    conv2d                        & (None, 192, 192, 32)        & 896                  \\ \hline
    conv2d\_1                      & (None, 192, 192, 16)        & 4624                 \\ \hline
    batch\_normalization          & (None, 192, 192, 16)        & 64                   \\ \hline
    dropout                       & (None, 192, 192, 16)        & 0                    \\ \hline
    max\_pooling2d                & (None, 96, 96, 16)          & 0                    \\ \hline
    conv2d\_2                      & (None, 96, 96, 64)          & 9280                 \\ \hline
    conv2d\_3                      & (None, 96, 96, 32)          & 18464                \\ \hline
    dropout\_1                     & (None, 96, 96, 32)          & 0                    \\ \hline
    batch\_normalization\_1        & (None, 96, 96, 32)          & 128                  \\ \hline
    max\_pooling2d\_1              & (None, 48, 48, 32)          & 0                    \\ \hline
    conv2d\_4                      & (None, 48, 48, 128)         & 36992                \\ \hline
    conv2d\_5                      & (None, 48, 48, 64)          & 73792                \\ \hline
    batch\_normalization\_2        & (None, 48, 48, 64)          & 256                  \\ \hline
    dropout\_2                     & (None, 48, 48, 64)          & 0                    \\ \hline
    max\_pooling2d\_2              & (None, 24, 24, 64)          & 0                    \\ \hline
    conv2d\_6                      & (None, 24, 24, 256)         & 147712               \\ \hline
    conv2d\_7                      & (None, 24, 24, 128)         & 295040               \\ \hline
    batch\_normalization\_3        & (None, 24, 24, 128)         & 512                  \\ \hline
    dropout\_3                     & (None, 24, 24, 128)         & 0                    \\ \hline
    max\_pooling2d\_3              & (None, 12, 12, 128)         & 0                    \\ \hline
    conv2d\_8                      & (None, 12, 12, 512)         & 590336               \\ \hline
    conv2d\_9                      & (None, 12, 12, 256)         & 1179904              \\ \hline
    batch\_normalization\_4        & (None, 12, 12, 256)         & 1024                 \\ \hline
    dropout\_4                    & (None, 12, 12, 256)         & 0                    \\ \hline

  \end{tabular}
   \caption{Layer Parameters of U-Net Encoder}
  \label{tab:layerencoder}
\end{table}

\begin{table}[hb!]
\centering
\begin{tabular} {|c|c|c|}
\hline
\textbf{Layer Name}           & \textbf{Output Shape}        & \textbf{Parameters} \\ \hline
conv2d\_transpose             & (None, 24, 24, 128)         & 131,200                       \\ \hline
concatenate                   & (None, 24, 24, 256)         & 0                             \\ \hline
conv2d\_10                     & (None, 24, 24, 256)         & 590,080                       \\ \hline
conv2d\_11                     & (None, 24, 24, 256)         & 590,080                       \\ \hline
batch\_normalization\_5        & (None, 24, 24, 256)         & 1,024                         \\ \hline
dropout\_5                    & (None, 24, 24, 256)         & 0                             \\ \hline
conv2d\_transpose\_1           & (None, 48, 48, 64)          & 65,600                        \\ \hline
concatenate\_1                & (None, 48, 48, 128)         & 0                             \\ \hline
conv2d\_12                     & (None, 48, 48, 128)         & 147,584                       \\ \hline
batch\_normalization\_6        & (None, 48, 48, 128)         & 512                           \\ \hline
dropout\_6                    & (None, 48, 48, 128)         & 0                             \\ \hline
conv2d\_13                     & (None, 48, 48, 128)         & 147,584                       \\ \hline
batch\_normalization\_7        & (None, 48, 48, 128)         & 512                           \\ \hline
add\_1                        & (None, 48, 48, 128)         & 0                             \\ \hline
conv2d\_transpose\_2           & (None, 96, 96, 32)          & 16,416                        \\ \hline
concatenate\_2                & (None, 96, 96, 64)          & 0                             \\ \hline
conv2d\_14                     & (None, 96, 96, 64)          & 36,928                        \\ \hline
batch\_normalization\_8        & (None, 96, 96, 64)          & 256                           \\ \hline
dropout\_7                    & (None, 96, 96, 64)          & 0                             \\ \hline
conv2d\_15                     & (None, 96, 96, 64)          & 36,928                        \\ \hline
batch\_normalization\_9        & (None, 96, 96, 64)          & 256                           \\ \hline
add\_2                        & (None, 96, 96, 64)          & 0                             \\ \hline
conv2d\_transpose\_3           & (None, 192, 192, 16)        & 4,112                         \\ \hline
concatenate\_3                & (None, 192, 192, 32)        & 0                             \\ \hline
conv2d\_16                     & (None, 192, 192, 32)        & 9,248                         \\ \hline
batch\_normalization\_10       & (None, 192, 192, 32)        & 128                           \\ \hline
dropout\_8                    & (None, 192, 192, 32)        & 0                             \\ \hline
conv2d\_17                     & (None, 192, 192, 32)        & 9,248                         \\ \hline
batch\_normalization\_11       & (None, 192, 192, 32)        & 128                           \\ \hline
add\_3                        & (None, 192, 192, 32)        & 0                             \\ \hline
conv2d\_18                     & (None, 192, 192, 3)         & 99                            \\ \hline
\end{tabular}

\caption{Layer Parameters of U-Net Decoder}
\label{tab:layerdecoder}
\end{table}

\begin{table}[htpb]
  \centering
  \begin{tabular}{|c|c|}
  \hline
  \textbf{Parameter Type} & \textbf{Parameters} \\
      \hline
   Total Params & 4,146,947 \\
    \hline
   Trainable Params & 4,144,547 \\
    \hline
    Non-trainable Params & 2,400 \\
\hline
  \end{tabular}
  
  \caption{Total Parameter Summary}
  \label{tab:params}
\end{table}

In order for the model to accurately segment input pictures, the suggested U-Net architecture is built on integrating encoder and decoder structures Figure \ref{fig:pu}. 

The encoder [Table \ref{tab:layerencoder}] uses Convolutional layers with a 3x3 kernel size and a large number of filters to downsample the input. To reverse the process and restore the original picture, the decoder uses transposed Convolutional layers to upsample the encoded information. A popular tool for picture segmentation, 3x3 Convolutional filters are trained in Deep Neural Networks using the Rectified Linear Unit (ReLU) activation function. Mixing batch normalization layers with dropout layers at rates ranging from 0.1 to 0.3 improves learning and avoids overfitting. A 2D Convolutional layer with 32 filters and a 3x3 kernel is the first layer in the encoder. Then comes another layer with 16 filters. After the two layers are normalized in batches, a 0.1-rate dropout layer is applied. A 2x2 max pooling layer then reduces the feature maps' spatial resolution. A max pooling layer, two Convolutional layers, a batch normalization layer, and a dropout layer make up each of the five encoder blocks.

Decoding is structurally identical to encoding [Table \ref{tab:layerdecoder}]; both use a concatenation layer as their foundation and incorporate two transposed Convolutional layers. With 128 filters in the first block and 8 in the final, the transposed Convolutional layers gradually reduce the filter count. For example, in later decoder blocks, the residual link between the concatenation layer and the first Convolutional layer improves gradient flow and the network's overall efficiency. Except for the last layer of each block, which does not use an activation function, all Convolutional layers use ReLU. The problem of Deep Neural Networks' disappearing gradients is solved by including residual connections. The overall parameter summary is listed down in Table \ref{tab:params}.

\subsection{Federated Learning(FL)}
\label{Federated Learning}
\begin{figure}[htbp]
\centerline{\includegraphics[width=1.2\textwidth]{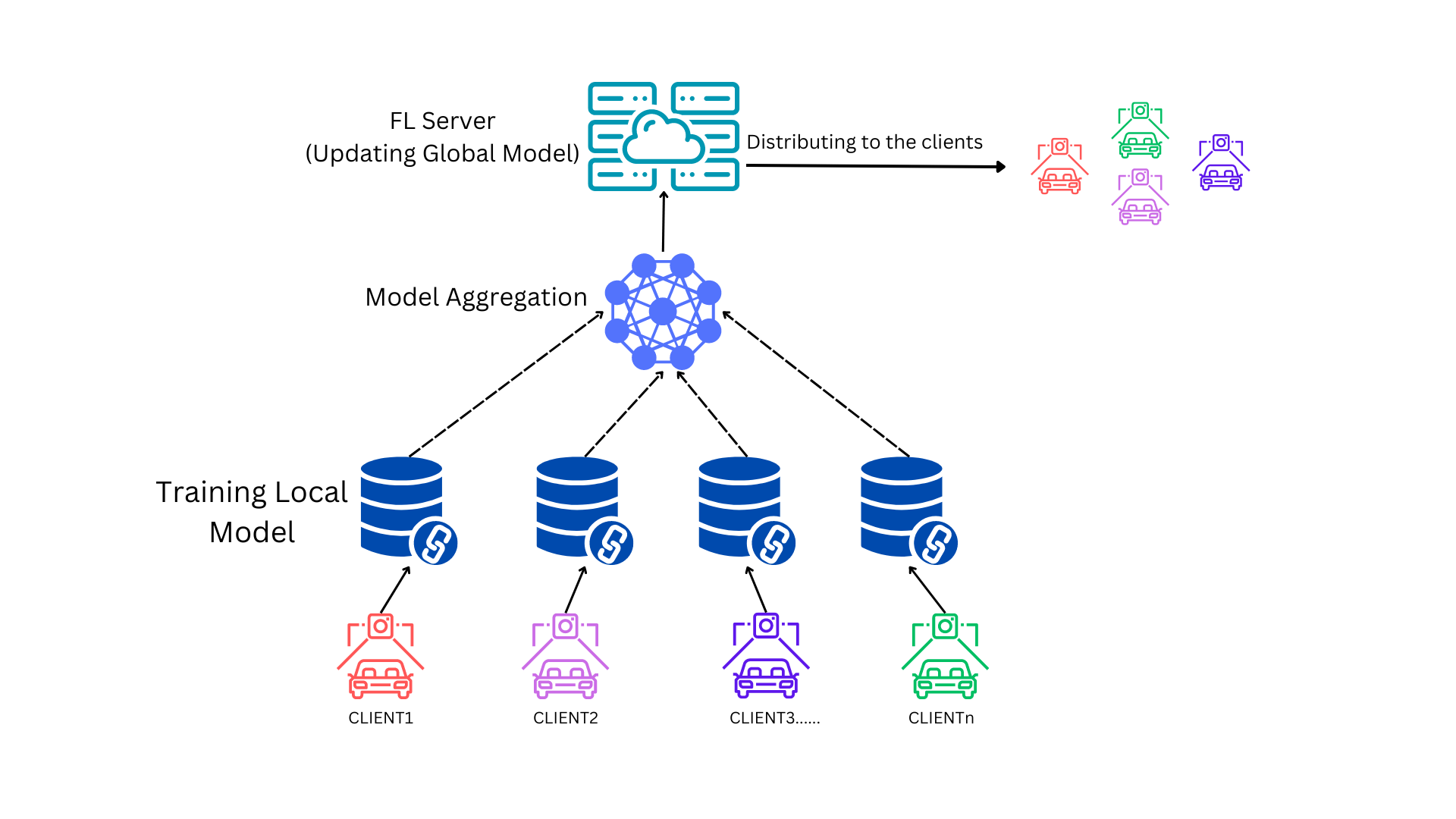}}
\vspace{4 mm}
\caption{Proposed FL Architecture}
\label{fig:fl}
\end{figure}

This model implements Federated Learning along with U-Net Segmentation, in which the FL model training is decentralized and the data remains on the devices of the clients. The U-Net and Federated Learning-Based Model take advantage of the privacy-maintaining factor of FL, along with the task of performing license plate detection on a vehicle image dataset. The main goal is to achieve the ability to distribute the training data among the clients while also maintaining the privacy of the users. First, by distributing training data across multiple clients by dividing the training data into separate sets, the sets represent a client. In order to start the data preparation equation \ref{eq:fl}, the total number of images in the training set is divided equally across the four clients. 
\begin{equation}
{Data\,Partition}= \frac{Total\,Number\,of\,Images}{Number\,of\,Clients}
 \label{eq:fl}
\end{equation}
Data preprocessing was done by reshaping the pixel data and labels, repeating the dataset for multiple epochs, shuffling the data using a specified buffer size, batching the data according to a predefined batch size, and mapping the data to a suitable format. Afterwards, federated data from the client-specific data and associated clients was created, which supports the preprocessing steps and also generates a list of preprocessed datasets for each client. In this model (Figure \ref{fig:fl}), four clients were used, where they were equally distributed datasets for training and validation purposes and used the same proposed U-Net model among the clients. The client then conducts local training using the initial model, and subsequently, the client calculates updates or gradients based on their locally trained model. To protect the privacy of the data, secure aggregation techniques were used to send the updates to the central server without affecting the raw data. Once the updates were collected, the updated model was sent back to the clients so that more iterations could be done.

\subsection{OCR}
\label{OCR}
\begin{figure}[htb]
\centerline{\includegraphics[width=\textwidth]{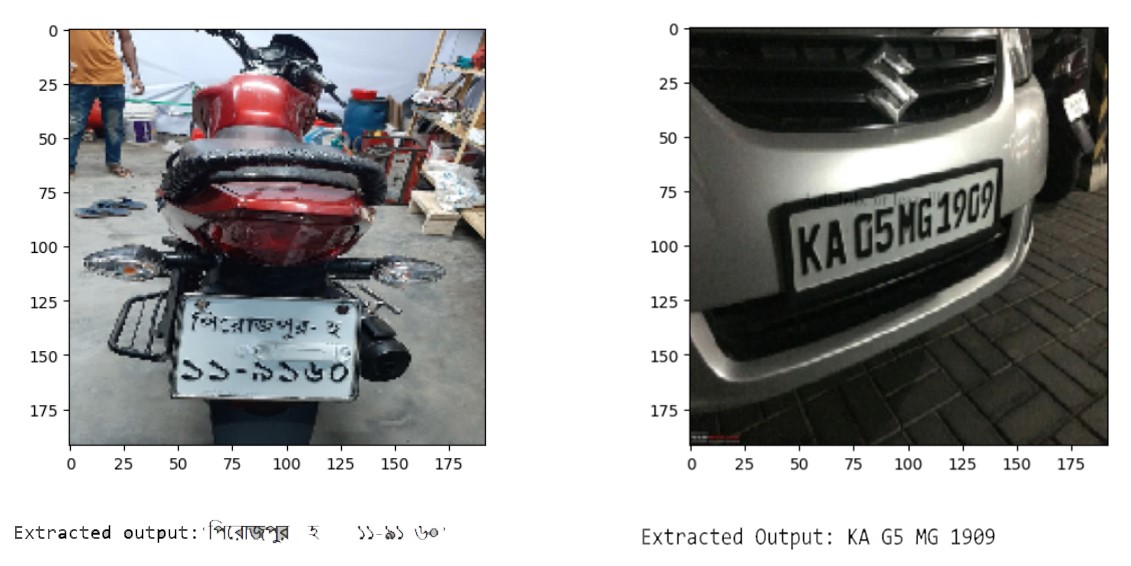}}
\caption{Extracted Vehicle Number Using OCR From FL U-Net Model Output}
\label{fig:oc1}
\end{figure}

OCR was used to extract license plate texts from the segmented images after completing the Federated U-Net model. The model of OCR being used here is Tesseract for both English and Bangla characters, which helps to achieve significant results and is also open source. Figure \ref{fig:oc1} is the result of OCR extracting Bangla and English characters from a given input image.

\section{Statistical Metrics}
\label{Statistical metrics}
The statistical measures used were Dice Score Coefficient (DSC), Cosine Similarity (CS), Binary Cross Entropy (BCE) Loss, Intersection over Union (IoU), Structural Similarity (SSIM), Root Mean Square Error (RMSE), Accuracy, Recall, Precision, F1 Score, and Area Under the ROC Curve (AUC).
\begin{itemize}
    \item 
    The Dice Score Coefficient (Equation \ref{eq:dsc}), which may range anywhere from 0 to 1, is used to assess the performance of models. A score of 1 indicates a pixel-perfect match between the ground truth annotation and the model’s output.

\begin{equation}
    DSC = \frac{2 \cdot \sum (p_{true} \cdot p_{pred}) }{ \sum p_{true} + \sum p_{pred} + \epsilon}
\label{eq:dsc}
\end{equation}

\item 
Cosine Similarity (Equation \ref{eq:sim}), is a metric that is used in many different machine learning algorithms to calculate the distance between neighbors. 

\begin{equation}
{Cosine\ Similarity}(A, B) = \frac{{cosine}(\theta)}{\|A\| \cdot \|B\|}
\label{eq:sim}
\end{equation}

\item 
The objective of the Binary Cross Entropy (BCE) Loss (Equation \ref{eq:BCE LOSS}), is to determine the degree to which the actual and anticipated picture masks vary in the amount of information that they contain. 

\begin{equation}
BCE(y, \hat{y}) = -[y \cdot \log(\hat{y}) + (1 - y) \cdot \log(1 - \hat{y})]
\label{eq:BCE LOSS}
\end{equation}

\item 
In the context of object identification and segmentation, Intersection over Union (Equation \ref{eq:iou}) evaluates the degree of overlap between the ground truth area and the prediction zone. 

\begin{equation}
IoU = \frac{{{TP}}}{{{TP + FP+ FN}}}
\label{eq:iou}
\end{equation}

\item 
The Root Mean Square Error (Equation \ref{eq:cos_sim}) is a performance metric for regression models. 

\begin{equation}
{RMSE} = \sqrt{\frac{1}{n}\sum_{i=1}^{n}(y_i - \hat{y}_i)^2}
\label{eq:cos_sim}
\end{equation}

\item 
Accuracy (Equation \ref{eq:acc}) alludes to the total number of instances where the model produced accurate predictions. 
	
\begin{equation}
{Accuracy} = \frac{{TP+TN}}{{TP + TN + FP+ FN}}
\label{eq:acc}
\end{equation}

\item 
Recall (Equation \ref{eq:rec}) is the percentage of data samples that a model correctly classifies as belonging to a class of interest (the “positive class”), relative to the total number of samples for that class. 

\begin{equation}
{Recall} = \frac{TP}{{TP} + {FN}}
\label{eq:rec}
\end{equation}

\item 
The level of Precision (Equation \ref{eq:Level}) is used to determine if the percentage of identifications is accurate. 

\begin{equation}
    {Precision} = \frac{TP}{{TP} + {FP}}
\label{eq:Level}
\end{equation}

\item 
The Structural Similarity (Equation \ref{eq:ssim}) is a perceptual metric that evaluates the decline in picture quality due to either data compression or data transmission losses. 

\begin{equation}
{SSIM}(x, y) = (l(x, y) \cdot c(x, y) \cdot s(x, y))
\label{eq:ssim}
\end{equation}

\item 
The factor F1 Score (Equation \ref{eq:f1}) is a classification model’s capacity to return results.

\begin{equation}
{F1\ Score} = \frac{2 \cdot (Precision \cdot Recall)}
{Precision + Recall}
\label{eq:f1}
\end{equation}

\item 
Area Under the ROC Curve (Equation \ref{eq:auc}) indicates the probability that a random positive example precedes a random negative example. AUC values range between 0 and 1. A model with 100\% incorrect predictions has an AUC of 0, while value 1 indicates perfect predictions.

\begin{equation}
AUC =\frac{\sum(FPR[i+1] - FPR[i]) \cdot (TPR[i] + TPR[i+1])}{2}
\label{eq:auc}
\end{equation}

\end{itemize}

\section{Quantitative Analysis}

\subsection{Aggregator Comparison}

Using numerical data, quantitative research tests hypotheses about the relationships between variables to see if there is a causal or correlational link. A performance assessment (Table \ref{tab:model_metrics FL}) was used to assess the aggregators' efficiency, which exposed unique features of the three aggregators. The Mean Factory scored the highest on several measures that indicate how well it matches or overlaps with the ground truth; those include IoU, SSIM, Summation of Cosine Distance, and Dice Coefficient. After looking at these measures side by side, it can be seen that the private quantile estimation process performed poorly when it came to similarity and overlap. With the smallest RMSE, the deferentially private factory showed that there was very little variation from the ground truth pixel-wise. On the other hand, the private quantile estimation process had the highest RMSE and BCE values when compared to the ground truth. This meant that the differences between pixels were the most inaccurate. Important for comprehending the aggregators' efficacy in the Federated Learning system, these results give a quantitative evaluation of their performances.

\begin{table}[htpb]
\begin{tabular}{| p{4.6 em}|p{3.5 em}|p{3.2 em}|p{3.1 em}|p{3.1em}|p{3.1em}|p{3.4em}|}
\hline
Aggregator & Dice Coefficient & BCE Dice Loss & IOU & RMSE & SSIM & SCD \\
\hline
Mean Factory & 0.871 & 0.84 & 0.781 & 0.0017& 0.014 &827.471 \\ \hline

PQEP & 0.7448& 0.7810 & 0.7246 & 0.028 & 0.2010 & 844.5041 \\  \hline

DPF & 0.8241 & 0.8054 & 0.7681 & 0.010 & 0.017 & 825.038 \\ \hline
\end{tabular}
\caption{Model Evaluation Metrics for FL Aggregators}
\label{tab:model_metrics FL}
\end{table}

\begin{table}[hb]
\centering
\begin{tabular}{| p{3.8em}|p{3 em}|p{3 em}|p{3 em}|p{3 em}|p{3 em}|p{4 em}|}
\hline
Model & DC & BCE Dice Loss & IOU & RMSE & SSIM & SCD \\
\hline
FL U-Net & 0.883 &	0.852 & 0.798 &	0.002 & 0.018 & 1049.7 \\ \hline
YOLO & 0.701&	0.826 & 0.715 & 0.0953 &	0.181 &	1101.6 \\ \hline
Semantic U-Net & 0.8735 & 0.8467& 0.798 & 0.002 &	 0.028 & 1050.6\\ \hline

\end{tabular}
\caption{Model Evaluation Metrics for the Merged Dataset}
\label{tab:model_metrics M}
\end{table}

\subsection{Model Comparison}
\label{model comparison}
\begin{figure}[htpb!]
\centerline{\includegraphics[width=\textwidth]{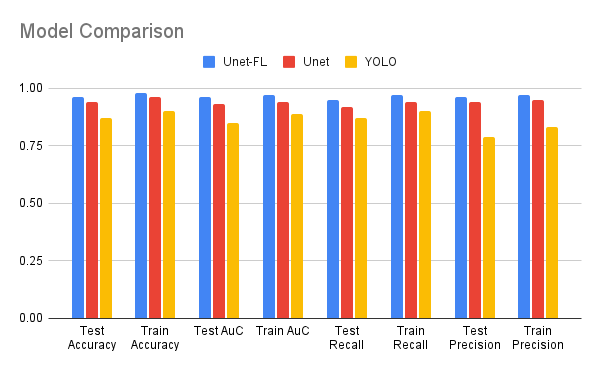}}
\vspace{2mm}
\caption{Model Comparison of FL U-Net, U-Net and YOLO}
\label{fig:model comp}
\end{figure}

From Table \ref{tab:model_metrics M}, it can be seen that for irregular masking, U-Net and FL U-Net perform substantially well. On the other hand, YOLO was not able to detect irregular masking properly. The true mask is the masking done manually and then compared to U-Net, FL U-Net, and YOLO. When compared to other models, Federated U-Net shows higher values for the DICE and IOU suggesting that it has better overlap with the ground truth masks. The models also exhibit low error and dissimilarity between predicted and ground truth masks, as they have the least BCE, RMSE, and Summation of Cosine Distance. SSIM results show that the bounding box YOLO model has the best structural similarity between the ground truth mask and the predicted one. Nevertheless, the lowest DICE and IOU indicate that it exhibits the weakest overlap. While Federated U-Net and Semantic U-Net both do well on the DICE and IOU metrics, Semantic U-Net is marginally behind on the BCE, RMSE, and Sum of Cosine Distance metrics. There is less structural similarity between the predicted and ground truth masks since their SSIM is lower than the bounding box YOLO. Among all the models tested, the Federated U-Net model proved to be the most reliable. In terms of both testing and training, Federated U-Net performed the best (Figure \ref{fig:model comp}) in the cases of Accuracy, AUC, Recall and Precision. U-Net was the second-best performer, and finally, YOLO came in last in terms of overall performance.

\section{Qualititive Analysis}
\label{qualitive analysis }
\begin{figure}[ht!]
\centerline{\includegraphics[width=1.3\textwidth]{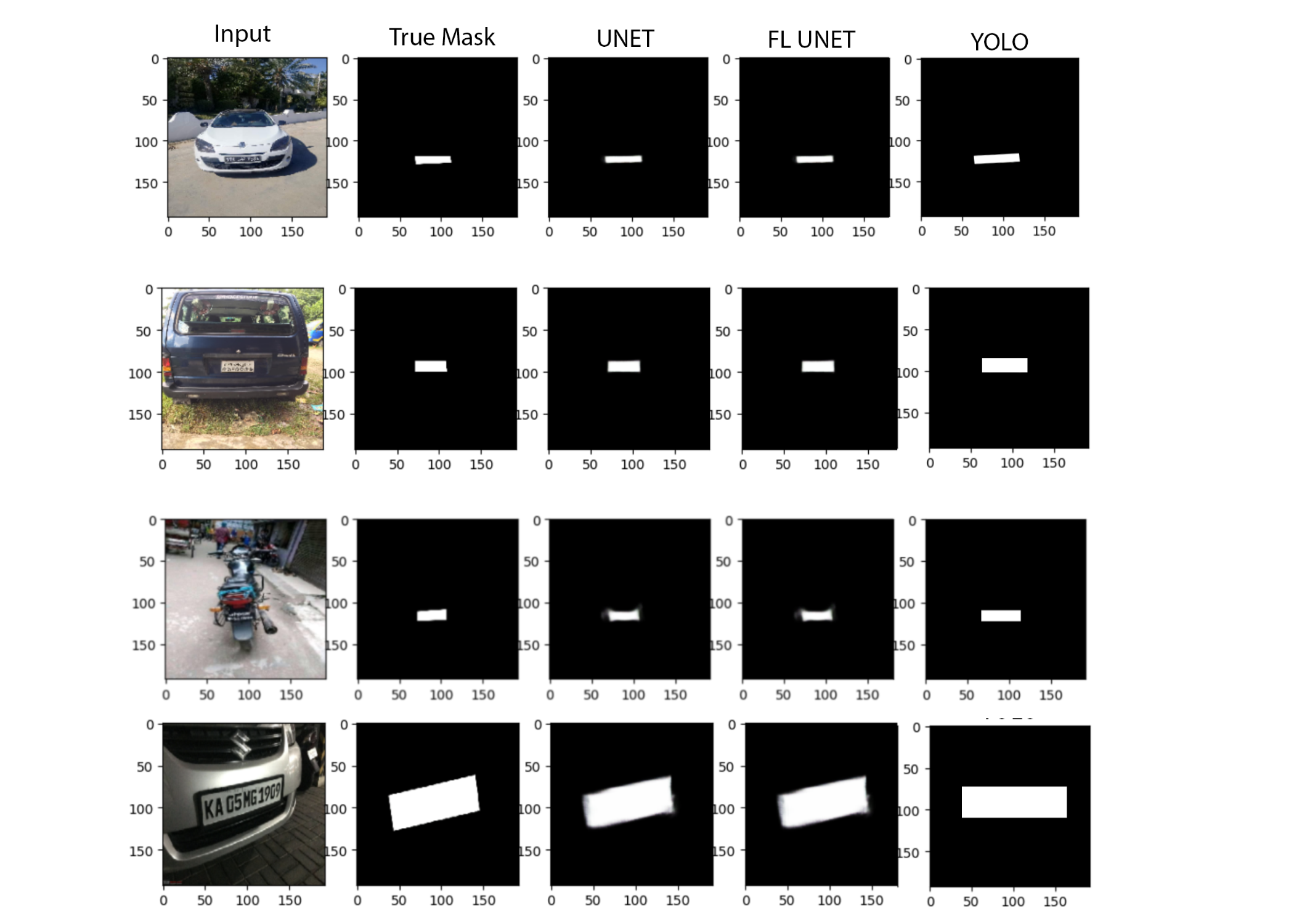}}
\caption{Qualitative Analysis of FL U-Net, Sementic U-Net and YOLO}
\label{fig:qsd M}
\end{figure}

When comparing the models (Figure \ref{fig:qsd M}) to the true mask of the fifth row image, which contains irregular masking, YOLO even performed flawlessly for the second image of regular license plate but not so well for the irregular license plate of fifth image. On the other hand, both sementic U-Net and FL U-Net were able to successfully complete the task by conducting both regular and irregular masking. In the proposed model, the irregular masking detected more precisely compared to other models.

\newpage
\section{Model Performance}

\begin{figure}[h!b]
    \includegraphics[width=.5\textwidth]{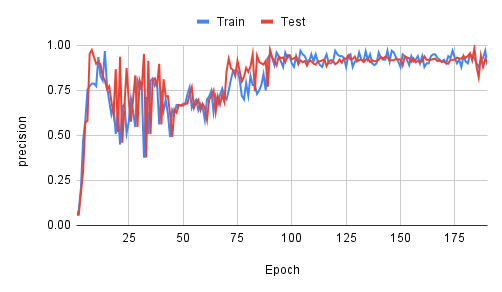}
    \includegraphics[width=.5\textwidth]{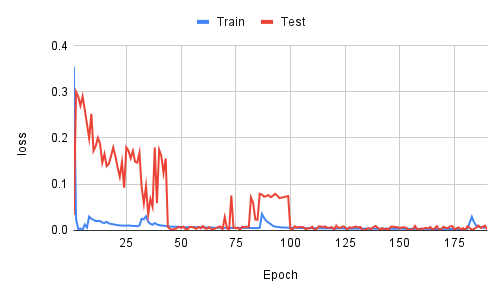}
    \includegraphics[width=.5\textwidth]{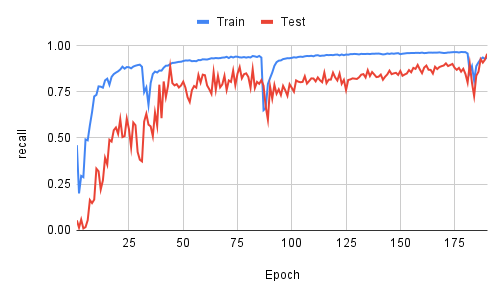}
    \includegraphics[width=.5\textwidth]{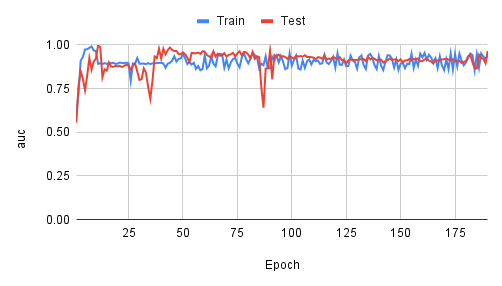}
    \begin{center}
    \includegraphics[width=.5\textwidth]{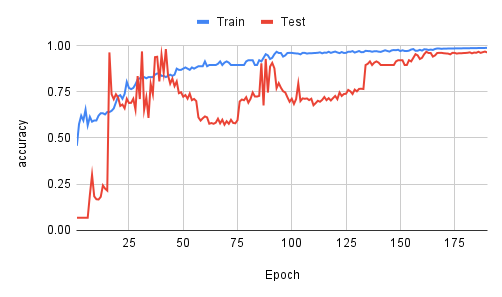}
    \end{center}

    \caption{Epoch-wise Precision, Loss, Recall, AUC and Accuracy for FL U-Net Model}
    \label{fig:result of fl U-Net}
\end{figure}

In the proposed FL U-Net, it can be observed that there is a rapid spike in accuracy at the beginning of training, which reaches its highest at 25 to 50 epochs and during testing at around 175 epochs and maintains a steady increase (Figure \ref{fig:result of fl U-Net}). During testing, the flow is not steady and suddenly decreases at different epochs. From the AUC figure \ref{fig:result of fl U-Net}, there has been a steady flow, but during testing, there has been a decrease around 50 epochs and the lowest at 85 epochs. During training, the loss at the start remains at its highest, and then afterwards, the loss goes down. Even in the testing stage, similar results can be seen, but from 75 to 100 epochs, there is a small increase, but it goes down. During training and testing, there remains a gradual instability where the precision fluctuates from 15 epochs to around 80 epochs and goes to its lowest at around 35 epochs. But around 100, the spike reaches a good amount of stability. The recall for testing and training increases gradually, and during training, the recall sometimes decreases, but the majority of the time it remains closer to the highest. During testing, the recall remains quite unstable from 0 to 100 epochs, but afterwards it reaches some stability.

\vspace{5mm}
In the proposed FL U-Net model, during training, FL U-Net got a score of 0.98 for Accuracy. For AUC, Recall, and Precision, there was a score of 0.97. In the case of loss, there remains a score of 0.0014 during training. On the other hand, during testing, a score of 0.96 for Accuracy, AUC and Precision were achieved. For Recall and Loss, a score of 0.95 and 0.0016 were found. Table \ref{tab:fu T} summarizes the results mentioned earlier. 

\begin{table}[htpb]
  \centering
 
  \begin{tabular}{|c|c|c|}
   \hline
   \textbf{Metric} &  \textbf{Train} & \textbf{Test}  \\ \hline
        
            Accuracy &  0.98  & 0.96\\
            \hline
            AUC &  0.97  & 0.96\\
            \hline
            Recall &  0.97  & 0.95\\
            \hline
            Loss &  0.0014 & 0.0016\\
            \hline
            Precision & 0.97  & 0.96\\

   \hline
  \end{tabular}
   \caption{Federated U-Net table }
  \label{tab:fu T}
\end{table}

\section{Conclusion}
To effectively monitor traffic, ensure correct vehicle licensing, identify cars participating in criminal activities, and solve issues like unauthorized parking, a license plate detection system is necessary in today's climate. There is an urgent need for an automatic license plate detection system in today's society since these activities cannot be performed manually and are thus prone to mistakes. Considering its durability compared to models depending on geometric bounding boxes and its ability to minimize irregular masking issues in poor settings, the inclusion of the U-Net model in this system is particularly remarkable, considering its relative rarity. Data privacy is further improved by using Federated Learning, which increases system dependability and security. In order to accomplish improving detection accuracy and benchmarking the model against current methods, more license plate photos can be gathered to increase the model's resilience and training breadth. Extending the system's license plate recognition capabilities to digital pictures is currently in the works, and online applications are being considered for the purpose of facilitating user access while protecting anonymity. To conclude with, this work intended to show how Federated Learning and U-Net may work together to build a system that is trustworthy, safe, and very resilient.


 \bibliographystyle{elsarticle-num} 
 \bibliography{PlateSegFL}
\end{document}